%% file: main_iccv.tex
\documentclass[10pt,twocolumn,letterpaper]{article}

\usepackage[final]{iccv}      

\input{preamble}

\usepackage{pifont}
\usepackage{float}
\usepackage{breqn}
\usepackage[normalem]{ulem} 
\usepackage[switch]{lineno}
\usepackage{array} 
\usepackage{multirow}
\usepackage{xcolor} 
\usepackage[many]{tcolorbox}
\usepackage{amsmath}
\usepackage{bbm}
\usepackage{balance}

\usepackage{booktabs}
\usepackage{rotating}
\usepackage{longtable}

\definecolor{iccvblue}{rgb}{0.21,0.49,0.74}
\usepackage[pagebackref,breaklinks,colorlinks,allcolors=iccvblue]{hyperref}

\definecolor{tagbg}{RGB}{224,234,241}   
\definecolor{tagtext}{RGB}{62,109,154}  

\definecolor{irrtagbg}{RGB}{255,220,220}  
\definecolor{irrtagtext}{RGB}{154,62,62}  
\usepackage{amsmath}
\DeclareMathOperator*{\argmax}{arg\,max}

\newcommand{\myirrtag}{\textcolor{irrtagtext}}
\newcommand{\mytag}{\textcolor{tagtext}}

\newcommand{\cmark}{\ding{51}}%
\newcommand{\xmark}{\ding{55}}%
\newcommand{\METHOD}{MAVias}

\def\paperID{8716} 
\def\confName{ICCV}
\def\confYear{2025}

\title{\METHOD: Mitigate any Visual Bias} 
\author{Ioannis Sarridis$^{1,2}$ ~~~~ Christos Koutlis$^{1}$ ~~~~ Symeon Papadopoulos$^1$   ~~~~
Christos Diou$^2$ \vspace{3pt} \\ 
$^1$Information Technologies Institute, CERTH, Greece\\ 
$^2$Department of Informatics and Telematics, Harokopio University of Athens, Greece\\ 
{\tt\small \{gsarridis, ckoutlis, papadop\}@iti.gr} ~~~~ {\tt\small \{isarridis, cdiou\}@hua.gr} 
}

\begin{document}

\twocolumn[{%
\renewcommand\twocolumn[1][]{#1}%
\maketitle
\begin{center}
    \centering
    \captionsetup{type=figure}
    \includegraphics[width=1\textwidth,trim={0.8cm 9.43cm 0.35cm 2.5cm},clip]{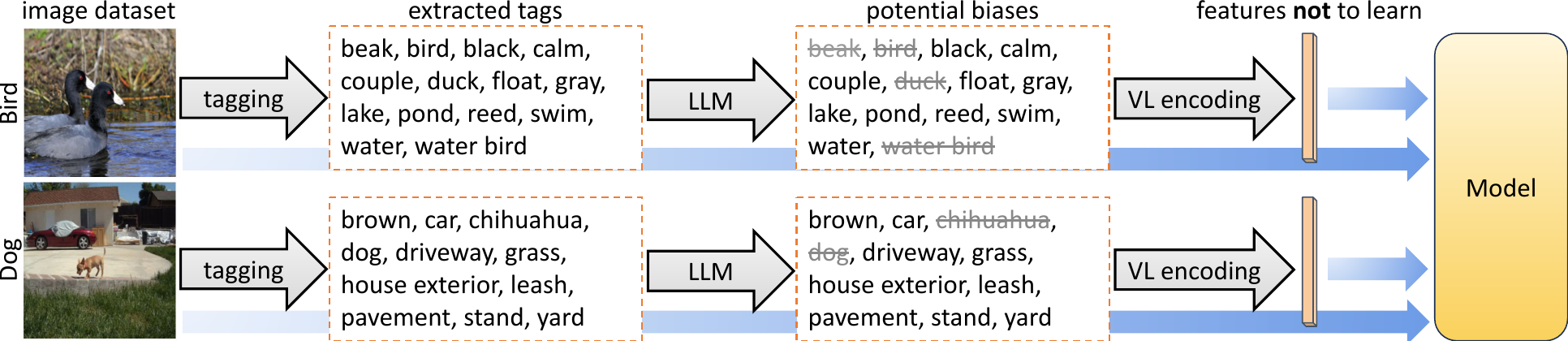}
    \captionof{figure}{    
    \METHOD\ identifies instance-level potential visual biases through foundational models that extract tags representing visual features and assess relevance to the target class. Then, \METHOD\ encodes these features within the vision-language space and integrates them into a bias-aware framework to train a model that is invariant to such visual biases.}
    \label{fig:teaser}
\end{center}%
}]
\begin{abstract}
Mitigating biases in computer vision models is an essential step towards trustworthy artificial intelligence systems. Existing bias mitigation methods are limited to predefined biases, preventing their use in visual datasets where multiple, possibly unknown biases exist. To address this limitation, we introduce \METHOD , an open-set bias mitigation approach that leverages foundation models to discover spurious associations between visual attributes and target classes. \METHOD\ first captures a wide variety of visual features in natural language via a foundation image tagging model, and then leverages a large language model to select visual features that define the target class, resulting in a set of language-coded potential visual biases. It then translates these biases into vision-language embeddings and introduces an in-processing bias mitigation approach to prevent the model from encoding information related to them. Experiments on diverse datasets, including CelebA, Waterbirds, ImageNet, and UrbanCars, show that \METHOD\ effectively detects and mitigates a wide range of biases in visual recognition tasks, outperforming current state-of-the-art. 
\end{abstract}
\vspace{-6mm}

\section{Introduction}
Computer Vision (CV) progress has been largely driven by Deep Learning (DL) advances \cite{dosovitskiy2020image,ho2020denoising} and large-scale datasets \cite{deng2009imagenet,lin2014microsoft,krishna2017visual}, enabling models to learn complex patterns and visual features with impressive accuracy. However, alongside this progress, concerns have emerged about biases embedded in these models \cite{ntoutsi2020bias,mehrabi2021survey, fabbrizzi2022survey,sarridis2023towards,sarridis2024facex,wang2022revise} -- often stemming from unintended correlations present in the training data \cite{liu2024devil, gupta2024bias, melnychuk2024bounds, li2024be}. The correlated attributes that are irrelevant act as ``shortcuts'' and can significantly impact the model's reliability and generalization \cite{ramos2025,sagawa2019distributionally,gupta2018robot}. 
It is important to note that in the context of this paper, we define as visual bias any characteristic that does not contribute to defining the target class, which we refer to as ``irrelevant''.
To address this, several methodologies have been developed. Broadly, these fall into two categories: Bias label-Aware (BA) and label-Unaware (BU) methods. BA methods leverage the annotations of attributes introducing the biases to address them \cite{kim2019lnl,wang2020DI,tartaglione2021end,barbano2022fairkl}. BU methods focus on extracting bias pseudo-labels in cases of extreme biases where a bias-proxy model (or bias-capturing classifier) can be trained using the task's target labels that closely align with the bias labels \cite{hong2021bb,nam2020LfF,sarridis2023flac,ludebiasing}.

While effective in certain contexts, both BA and BU methods have limited applicability when there are multiple, complex, and possibly unknown biases. Common challenging scenarios include:
\begin{itemize}
    \item 
    \textbf{Unknown biases}: In large general-purpose CV datasets, such as ImageNet, biases are difficult to identify and largely remain unknown, as they vary widely across different classes and are not prominent enough to allow training of bias proxy models. For instance, in the ImageNet9 example of Tab.~\ref{tab:sample_tags}, a sample labeled as a \textit{dog} could introduce biases related to the background scene (e.g., \textit{armchair}, \textit{couch}, \textit{pillow}, \textit{red}), the color of the dog (e.g., \textit{black} and \textit{white}), or accessories like a \textit{neckband}.
    \item 
    \textbf{Potential biases beyond a predefined set}: The CelebA example in Tab.~\ref{tab:sample_tags}, shows that beyond the \textit{hair color}, additional biases may be present, such as clothing styles (e.g., \textit{business suit}, \textit{tie}).
    \item 
    \textbf{Poor representation of predefined bias}: In some cases, biases are reduced to single labels, such as ``rural background'' in the UrbanCars dataset. However, as shown in Tab.~\ref{tab:sample_tags}, more nuanced descriptors at the instance level (e.g., \textit{path}, \textit{tree}, \textit{wood}, \textit{forest}, \textit{hydrant}, \textit{red}, etc.) provide a richer representation of bias.
\end{itemize}
\begin{table*}[t]
\centering
\caption{Examples of extracted tags for various datasets. Red color indicates the irrelevant tags ($\mathcal{B}^{(i)}$).}
\label{tab:sample_tags}
\footnotesize

\resizebox{\textwidth}{!}{%
\begin{tabular}{p{2.cm}p{3cm}p{3.6cm}p{3.6cm}p{3cm}}
\toprule
Dataset & ImageNet9 & Waterbirds  & UrbanCars  & CelebA  \\ \midrule
Target Class & dog  & bird species & car type & hair color \\ \midrule
Bias Type & unknown & background 
& background and objects 
& gender \\ \midrule
Sample & \includegraphics[width=3.0cm,height=3.0cm,trim={2.5cm 0cm 2.5cm 0},clip]{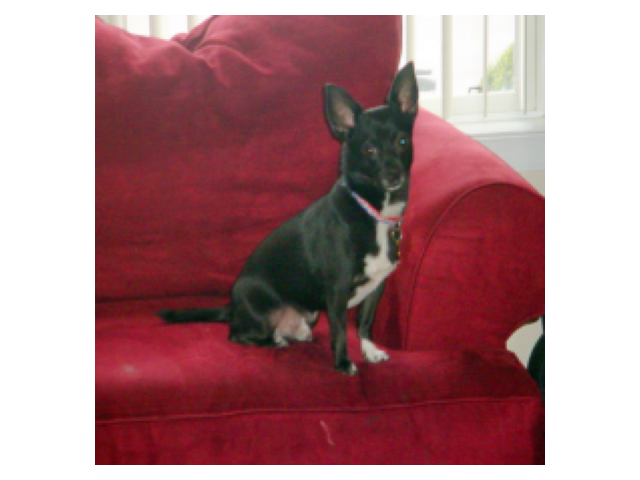} & \includegraphics[width=3.0cm,height=3.0cm,trim={2.5cm 0cm 2.5cm 0},clip]{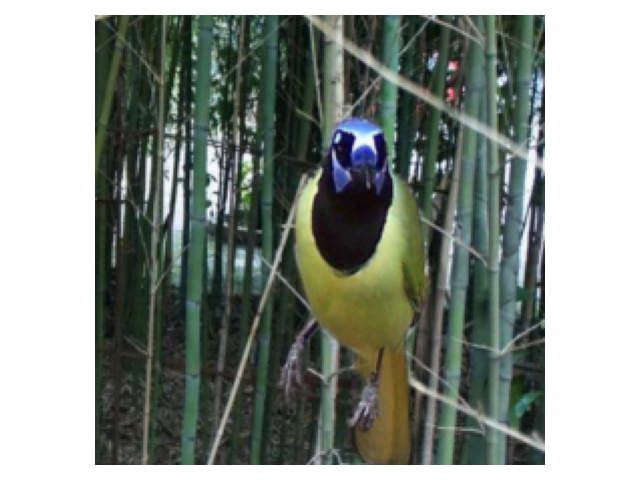} & \includegraphics[width=3.0cm,height=3.0cm,trim={2.5cm 0cm 2.5cm 0},clip]{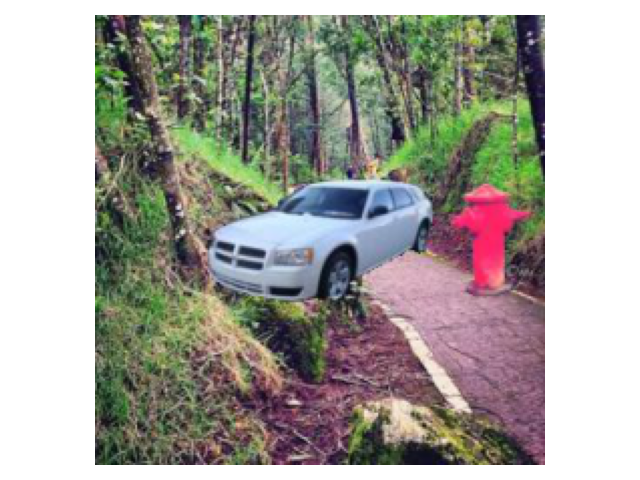} & \includegraphics[width=3.0cm,height=3.0cm,trim={2.5cm 0cm 2.5cm 0},clip]{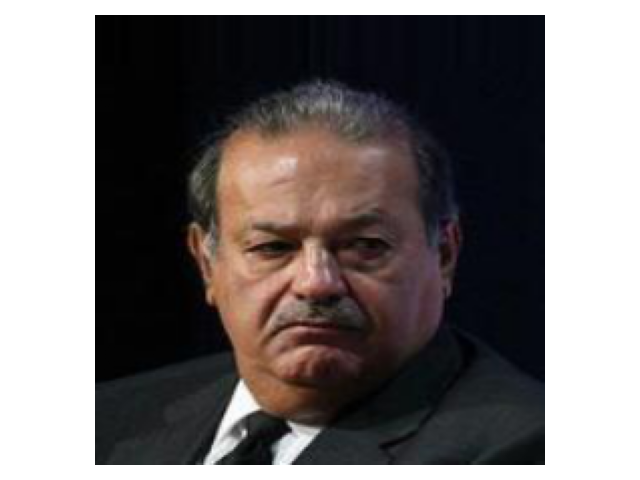} \\ \midrule
Extracted Tags & \myirrtag{armchair}, \myirrtag{black}, \myirrtag{chair}, \myirrtag{couch}, \mytag{dog}, \myirrtag{neckband}, \myirrtag{pillow}, \myirrtag{red}, \myirrtag{sit}, \myirrtag{white} & \myirrtag{bamboo}, \myirrtag{bamboo forest}, \mytag{bird}, \myirrtag{blue}, \myirrtag{branch}, \myirrtag{green}, \myirrtag{hide}, \mytag{parrot}, \myirrtag{perch}, \myirrtag{sit}, \myirrtag{stand}, \myirrtag{stem}, \myirrtag{tree}, \myirrtag{yellow} & \mytag{car}, \myirrtag{path}, \myirrtag{forest}, \myirrtag{hydrant}, \myirrtag{lush}, \myirrtag{park}, \myirrtag{red}, \myirrtag{road}, \mytag{sedan}, \myirrtag{silver}, \mytag{SUV}, \myirrtag{tree}, \myirrtag{white}, \myirrtag{wood} & 
\mytag{black}, \myirrtag{business suit}, \myirrtag{dress}, \myirrtag{dress shirt}, \myirrtag{man}, \myirrtag{stand}, \myirrtag{stare}, \myirrtag{suit}, \myirrtag{tie}, \myirrtag{wear}
\\ \bottomrule

\end{tabular}%
 }
\end{table*}

Existing BA and BU methods are not designed to mitigate such biases, leading to models that are biased and/or do not achieve their optimal generalization potential.
To address this, Mitigate Any Visual bias (\METHOD )  introduces a flexible and scalable solution, capable of identifying and mitigating open-set biases in CV datasets.
\METHOD\ begins by extracting descriptive tags that capture various visual features, such as general-purpose objects, actions, scenes, and visual attributes. 
Recent advancements in image tagging \cite{zhang2024recognize} cover large and comprehensive vocabularies (i.e., $>4,000$ tags), effectively meeting the open-set requirements of \METHOD.
Then, these tags are processed by a Large Language Model (LLM) to identify which ones are irrelevant to each of the target classes, leading to a rich set of \textit{language-encoded visual biases}, text descriptions of visual characteristics that are irrelevant to the classification task at hand.  \METHOD\ translates these biases into vision-language embeddings, projects them 
to the main model's backbone space and then to the classification layer. This projection layer is trained simultaneously alongside the main model, and during training, the output logits are a linear combination of the logits of the main model and those of the projection layer that captures visual biases.
This setup allows the main model to be exposed to the biased features -- those representing irrelevant information to the target class -- in a controlled way that leads to bias-invariant representations. Overall, \METHOD\ provides an effective, end-to-end solution for identifying and mitigating biases in open-set scenarios.
We evaluate the proposed method on several datasets involving single-attribute biases (CelebA, Waterbirds) as well as multi-attribute predefined biases (UrbanCars), demonstrating state-of-the-art performance. Furthermore, experiments were conducted on ImageNet9, involving unknown biases, where the suggested approach demonstrates significant gains (from 5.24\% to 10.21\%) in terms of accuracy compared to existing competitive approaches. 
\looseness=-1

The main contributions of this paper are the following: 
\begin{itemize}
\item A framework for identifying instance-specific open-set potential visual biases in  CV datasets.
\item A learning strategy that exploits foundation models to learn bias-invariant representations that force the under-training model to avoid encoding any number of identified potential biases per training sample.
\item An extensive evaluation study including 4 thematically diverse datasets demonstrating the effectiveness and general applicability of \METHOD , which outperforms the state-of-the-art.
\end{itemize}
\METHOD\ implementation is available as part of the VB-Mitigator library \cite{sarridis2025vbmitigator}.

\section{Related Work}
\paragraph{Bias identification.}
Several recent methods leverage text (such as captions, keywords, or tags) for bias detection, highlighting the potential of this approach in this domain. 
For instance, Say My Name (SaMyNa) \cite{ciranni2024say} is an explainability method that tries to discover model biases through text-based pipelines. 
Similarly, the Bias-to-Text (B2T) \cite{kim2024keywords} and Language-guided (Lg) \cite{zhao2024language} frameworks discover biases by extracting common keywords from the captions of misclassified images. These methods aim to identify the main source of bias rather than discovering potentially biased visual characteristics in an open-set setting.
Furthermore, OpenBias \cite{d2024openbias} is a framework for detecting biases in text-to-image generative models by leveraging LLMs to propose potential biases from captions. 
Although operating in a different domain (text-to-image generation), similarly to \METHOD\, OpenBias acknowledges the need for discovering biases in an open-set setting. However, it provides an LLM with text-only image descriptions asking for potential types of bias thus missing visual grounding of the depicted information, which could offer essential context and semantics. 
In contrast, \METHOD\ takes a more structured and systematic approach; it evaluates each descriptive tag by querying the LLM to determine whether it is directly relevant to the target class using a detailed prompt (specified in the supp. material). 

Furthermore, bias identification methods typically use a vanilla model and validation data to infer specific biases, while MAVias focuses on defining irrelevant visual features \emph{a priori}, and leveraging them in model training to mitigate them. While some open-set identification methods can provide bias labels to define subgroups for use by existing bias mitigation approaches \cite{ciranni2024say}, it is known that mitigation methods struggle with multi-attribute biases and cannot scale beyond single-attribute subgrouping \cite{li2023whac}. In contrast, MAVias uses instance-level irrelevant visual features without relying on dataset statistics (i.e., neither bias labels nor subgroups w.r.t. them are defined), allowing it to handle complex biases effectively.

\paragraph{Bias mitigation.}
Recent bias mitigation methods include those with direct access to the labels of attributes introducing bias (i.e., BA methods) \cite{kim2019lnl,wang2020DI,tartaglione2021end,hong2021bb,barbano2022fairkl,sagawa2019distributionally,qiu2023simple,li2023whac} and those that do not take advantage of such labels but instead rely on deriving pseudo-labels (i.e., BU methods) \cite{cadene2019rubi,bahng2020rebias,nam2020LfF, sarridis2023flac, wu2023discover, yang2024identifying, venkataramani2024causal}. 
For instance, Learning Not to Learn (LNL) \citep{kim2019lnl} is a BA method that discourages the model from predicting the attribute introducing bias, while
Bias Contrastive-Bias Balance (BC-BB) \citep{hong2021bb} and FairKL \citep{barbano2022fairkl} rely on the bias labels to enforce bias-neutral pairwise similarities between the samples using contrastive learning. Some methods have indirect access to the bias labels by utilizing bias-capturing classifiers trained on different datasets offering bias-related information explicitly \cite{sarridis2023flac,sarridis2024badd}. Finally, other methods infer pseudo-labels from the biased vanilla model to identify the biases \cite{nam2020LfF,bahng2020rebias, clark2019LM}. 
While these methods have been effective in mitigating biases, they primarily depend on predefined bias labels or pseudo-labels derived from biased models. 
On the contrary, this work explores open-set scenarios and introduces a flexible bias mitigation approach that can discover and handle multiple, diverse biases without requiring a bias-labeling system. 

\section{Methodology}
\label{sec:method}
\subsection{Problem Formulation}
Let $\mathcal{D} = \{(\mathbf{x}^{(i)}, y^{(i)})\}_{i=1}^{N}$ be a dataset consisting of $N$ images, where $\mathbf{x}^{(i)} \in \mathcal{X}$ represents an input image, and $y^{(i)} \in \mathcal{Y}$ is the corresponding target label. The goal is to train a DL model $f_{\boldsymbol{\theta}}: \mathcal{X} \to \mathcal{Y}$, parameterized by $\boldsymbol{\theta}$, to predict the target label $y^{(i)}$ given an image $\mathbf{x}^{(i)}$, while mitigating a set of potential biases, $\mathcal{B}^{(i)}$, present in $\mathbf{x}^{(i)}$ that may lead to biased predictions. In this context, the term ``potential biases'' refers to all visual attributes present in $\mathcal{X}$ that are irrelevant to the target class, such as elements of the background of images or other so-called ``shortcuts''.

\subsection{Method}

\subsubsection{Language-driven Bias Modeling}

In many general-purpose CV datasets, biases manifest through visual information that can be described in text. \METHOD, first, utilizes an image tagging model to extract tags that describe the visual information present in an image. Formally, for each image $\mathbf{x}^{(i)}$, we derive a set of tags $\mathcal{T}^{(i)} = \{t^{(i)}_{1}, t^{(i)}_{2}, \dots, t^{(i)}_{m_i}\}$, where $m_i$ represents the total number of tags for $i$-th sample. These capture various visual attributes, including colors, objects, backgrounds, and other features that can either describe the target class or potentially introduce bias.

The next step is to filter out tags that should not influence the decisions of an unbiased classifier. To achieve this, we leverage an LLM to identify which tags are irrelevant to the target class $y^{(i)}$. We denote this subset of potential biases as $\mathcal{B}^{(i)} \subseteq \mathcal{T}^{(i)}$. These encapsulate visual features that could lead to biased predictions if considered by the model. To ensure the LLM correctly identifies these irrelevant tags, we carefully design the prompt used for this task by providing precise instructions on what tag should be considered relevant (e.g., physical components, defining features, inherent characteristics, etc.) or not (e.g., background details, lighting, textures, other objects, etc.). Details on the prompt formulation process are provided in the supp. material. Table~\ref{tab:sample_tags} shows several examples of $\mathcal{B}^{(i)}$ for samples belonging to different datasets.

For each set of irrelevant tags $\mathcal{B}^{(i)}$ associated with an image $\mathbf{x}^{(i)}$, we employ a vision-language model to generate a single embedding $\mathbf{e}^{(i)} \in \mathbb{R}^d$, where $d$ is the dimension of the embedding. This is produced using the prompt ``a photo of $t^{(i)}_{1}, t^{(i)}_{2}, \dots, t^{(i)}_{k_i}$'', where $k_i \leq m_i$ is the number of irrelevant tags for the $i$-th sample. This aggregate embedding captures the combined information from all irrelevant tags, providing an instance-level representation of the biased features that could affect the model’s behavior. Comparison with alternative embedding approaches is provided in Sec.~\ref{sec:abl}.

\begin{figure*}[t]
    \centering
    \includegraphics[width=1.\textwidth,trim={0.1cm 0cm 2.33cm 0},clip]{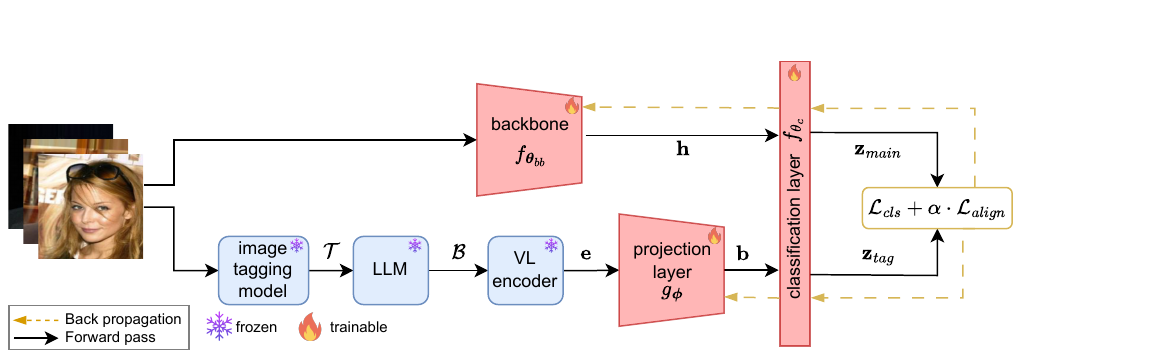}
    \caption{Illustration of the proposed framework for mitigating any visual bias during model training. For inference, only the backbone and the classification layer are considered (i.e., $f_{\boldsymbol{\theta}}$).}
    \label{fig:framework}
\end{figure*}
\subsubsection{Bias Mitigation}


We define the main model $f_{\boldsymbol{\theta}}(\mathbf{x}^{(i)})$, a DL classifier composed of the following:
(i) a backbone $f_{\boldsymbol{\theta}_{bb}}(\mathbf{x}^{(i)})$, which extracts feature representations $\mathbf{h}^{(i)}\in\mathbb{R}^r$, where $r$ is the feature vector size;
(ii) a classification head $f_{\boldsymbol{\theta}_{c}}(\mathbf{x}^{(i)})$ outputting the logits $\mathbf{z}^{(i)}_{\text{main}} = f_{\boldsymbol{\theta}_{c}}(\mathbf{h}^{(i)})\in\mathbb{R}^{p}$, where $p=\lvert\mathcal{Y}\rvert$.
The overall main model is expressed as
    $f_{\boldsymbol{\theta}}(\mathbf{x}^{(i)}) = f_{\boldsymbol{\theta}_{c}}(f_{\boldsymbol{\theta}_{bb}}(\mathbf{x}^{(i)}))$.
In parallel, we introduce a projection layer $g_{\boldsymbol{\phi}}$, parameterized by $\boldsymbol{\phi}$, which takes the visual bias embeddings $\mathbf{e}^{(i)}$ as input and outputs  embeddings $\mathbf{b}^{(i)}\in\mathbb{R}^p$. Note that $g_{\boldsymbol{\phi}}$ is employed to project $\mathbf{e}^{(i)}$ to the feature space of $\mathbf{h}^{(i)}$. Then the corresponding logits are derived through  $\mathbf{z}^{(i)}_{\text{tag}} = f_{\boldsymbol{\theta}_{c}}(\mathbf{b}^{(i)})\in\mathbb{R}^{p}$. 
The final logits $\mathbf{z}^{(i)}$ for each sample are the addition of the main model logits $\mathbf{z}^{(i)}_{\text{main}}$ and the visual bias logits $\mathbf{z}^{(i)}_{\text{tag}}$:
\begin{equation}
    \mathbf{z}^{(i)} = \mathbf{z}^{(i)}_{\text{main}} + \mathbf{z}^{(i)}_{\text{tag}}.
    \label{eq:logits_addition}
\end{equation}

It is worth noting that bias often arises during DL model training because biased attributes in the training data are easier to learn and thus dominate gradient updates \cite{pezeshki2021gradient}. To counteract this, \METHOD\ incorporates visual bias logits into the main model’s logits, ensuring that as the bias in a sample increases, its impact on gradient updates is reduced. The intuition behind this mechanism is that for bias-aligned samples, the value of $\mathbf{z}_{tag}$ is high, effectively reducing the magnitude of $\mathbf{z}_{main}$ and its contribution to the total logits $\mathbf{z}$. This leads to significantly reduced gradients for these samples. This is supported both empirically in Sec.~\ref{sec:results} and theoretically in the supp. material.
In other words, $\mathbf{z}_{tag}$ assists in decoupling the learning of biased features from the actual task at hand, which allows the main model to focus on the relevant features for the target prediction.

Furthermore, since both the main model and the projection layer are trained concurrently, it is essential to ensure the stability of the training process. To achieve this, we introduce a loss function that combines the classification loss with a logit alignment term. The classification loss ensures that the combined logits $\mathbf{z}^{(i)}$
accurately predict the target label $y^{(i)}$, while the alignment term controls the relative magnitudes of the main model’s logits $ \mathbf{z}^{(i)}_{\text{main}}$ and the visual bias logits $\mathbf{z}^{(i)}_{\text{tag}}$. 
By doing so, we prevent $g_{\boldsymbol{\phi}}$ from dominating or being overshadowed by $h_{\boldsymbol{\theta}}$.
Specifically, the loss is computed as:
   $\mathcal{L}(\boldsymbol{\theta}, \boldsymbol{\phi}) =   \mathcal{L}_{cls}(\mathbf{z}^{(i)}, y^{(i)}) + \alpha \cdot \mathcal{L}_{align}(\mathbf{z}^{(i)}_{\text{main}}, \mathbf{z}^{(i)}_{\text{tag}}) $,
   where:
   (i) \( \mathcal{L}_{cls}(\mathbf{z}^{(i)}, y^{(i)})\) is the classification loss (e.g., cross-entropy loss) between the final logits \(\mathbf{z}^{(i)}\) and the ground truth label \(y^{(i)}\); (ii)  $\mathcal{L}_{align}(\mathbf{z}^{(i)}_{\text{main}}, \mathbf{z}^{(i)}_{\text{tag}})$ is the logit alignment term for the norm of the logits, calculated as:
\begin{equation}
   \mathcal{L}^{(i)}_{align} = \frac{1}{2} \left\| \| \mathbf{z}^{(i)}_{\text{main}} \| - \lambda \cdot \| \mathbf{z}^{(i)}_{\text{tag}} \| \right\|^2 
   \label{eq:loss_reg}
\end{equation}
where $\|\cdot\|$ denotes the $l^2$-norm, $\lambda \in (0,1)$ is a scaling factor and $\alpha \in (0,1)$ is a weighting factor that balances the influence of the $\mathcal{L}_{align}(\mathbf{z}^{(i)}_{\text{main}}, \mathbf{z}^{(i)}_{\text{tag}})$ in the total loss function. Typically, the greater the bias in the data, the smaller the value of $\lambda$ should be, as this leads to smaller $\mathbf{z}_{main}$ values and reduced gradient updates for the bias-aligned samples. Details on the effects of hyperparameters $\lambda$ and $\alpha$ are provided in the supp. material. The overview of the proposed framework is illustrated in Fig.~\ref{fig:framework}.
\section{Experiments}
\subsection{Datasets}
We use Biased-CelebA \cite{hong2021bb}, Waterbirds \cite{sagawa2019distributionally}, UrbanCars \cite{li2023whac}, and Imagenet9 \cite{xiao2021noise}. Biased-CelebA is a subset of the CelebA dataset, containing facial images annotated with 40 binary attributes. In this subset, \textit{BlondHair} is the target attribute, while the \textit{gender} attribute introduces bias with a 90\% correlation. Waterbirds features a 95\% co-occurrence between waterbirds (or landbirds) and aquatic (or terrestrial) backgrounds. UrbanCars is an artificially generated dataset with a 95\% co-occurrence between car body types and relevant urban or rural backgrounds. ImageNet9 consists of 9 coarse ImageNet classes.\\
\subsection{Model Architectures}
For comparability purposes, we employ the same network architectures as those used in other works \cite{sarridis2023flac,hong2021bb,barbano2022fairkl,sarridis2024badd}. In particular, for CelebA, we adopt the ResNet-18 architecture \cite{he2016deep}, for Waterbirds, UrbanCars, and ImageNet9 datasets, we use ResNet-50 networks.
In all experiments, the projection layer is a dense layer that gets vision-language embeddings as input and its output size is aligned with the feature size of the main model. 
\subsection{Implementation Details}
The SGD optimizer is employed for all datasets except for CelebA, where Adam optimizer is used. We use an initial learning rate of 0.001, which is divided by 10 every 1/3 of the training epochs. The weight decay is set to $10^{-4}$. The batch size is 128 for CelebA and 64 for Waterbirds, UrbanCars, and ImageNet9.
Following previous works \cite{hong2021bb,sagawa2019distributionally,li2023whac}, we train the models for 40, 100, 300, and 40 for CelebA, Waterbirds, UrbanCars, and ImageNet9, respectively.
For Waterbirds and UrbanCars, we do not use a learning rate scheduler. The parameters $(\alpha, \lambda)$ are (0.01, 0.5), (0.05, 0.6), (0.01, 0.4), and (0.001, 0.7) for CelebA, Waterbirds, UrbanCars, and ImageNet9, respectively. RAM \cite{zhang2024recognize}, GPT-4o \cite{openai2023chatgpt4}, OpenCLIP \cite{clip} are employed for image tagging, irrelevant tag filtering, and vision-language encoding, respectively. 
For all the experiments presented in the main manuscript, we run the compared methods using the code provided by the corresponding papers, using the suggested hyperparameters. Specifically, we use 50 epochs for bias discovery and 100 upweight for JTT; $\alpha=110$ for FLAC-B; 0.1 SD coefficient; and $\alpha=0.7$ for LfF. For the baseline methods results presented in the supplementary material (i.e., closed-set), we present performance as reported in the corresponding works to avoid potential underperformance due to reimplementation.
Experiments were conducted on an NVIDIA A100 GPU. All experiments were repeated for 5 different random seeds.
\begin{table}[b]
    \centering
    \caption{Open-set performance comparison across CelebA, Waterbirds, and UrbanCars datasets.}
    \label{tab:openset}
    \begin{tabular}{clcc}
        \toprule
        Dataset & Method & WG Acc. & Avg. Acc.\\
        \midrule
        \multirow{7}{*}{\rotatebox{90}{CelebA}}  

            & LfF \cite{nam2020LfF}     & 14.7 \scriptsize{$\pm$15.2} & 67.1 \scriptsize{$\pm$4.4}  \\
            & JTT  \cite{liu2021just}   & \underline{31.5} \scriptsize{$\pm$8.0}  & 61.6 \scriptsize{$\pm$8.5}  \\
            & SD \cite{pezeshki2021gradient}     & 13.3 \scriptsize{$\pm$8.2}  & \underline{67.4 }\scriptsize{$\pm$1.9}  \\
            & Debian \cite{li2022discover}  & 12.0 \scriptsize{$\pm$8.7}  & 67.0 \scriptsize{$\pm$2.6}  \\
            & FLAC-B \cite{sarridis2023flac}  & 12.0 \scriptsize{$\pm$8.7}  & 65.9 \scriptsize{$\pm$2.4}  \\
            & \METHOD  & \textbf{66.7} \scriptsize{$\pm$4.7}  & \textbf{81.4} \scriptsize{$\pm$1.8}  \\
        \midrule
        \multirow{7}{*}{\rotatebox{90}{Waterbirds}}
            & LfF \cite{nam2020LfF}    & 30.0 \scriptsize{$\pm$6.8}  & 72.7 \scriptsize{$\pm$1.4}  \\
            & JTT  \cite{liu2021just}   & \underline{64.7} \scriptsize{$\pm$2.4}   & \underline{85.2} \scriptsize{$\pm$4.6}  \\
            & SD   \cite{pezeshki2021gradient}     & 35.0 \scriptsize{$\pm$16.3} & 75.5 \scriptsize{$\pm$4.1}  \\
            & Debian \cite{li2022discover} & 37.5 \scriptsize{$\pm$0.0}  & 74.7 \scriptsize{$\pm$0.5}  \\
            & FLAC-B  \cite{sarridis2023flac} & 37.5 \scriptsize{$\pm$8.8}  & 75.2 \scriptsize{$\pm$2.7}  \\
            & \METHOD  & \textbf{75.4} \scriptsize{$\pm$0.9}  & \textbf{87.5} \scriptsize{$\pm$1.2}  \\
                    \midrule
            \multirow{7}{*}{\rotatebox{90}{UrbanCars}}  
            & LfF \cite{nam2020LfF}     & 34.6 \scriptsize{$\pm$2.6}  & 61.0 \scriptsize{$\pm$1.4}  \\
            & JTT \cite{liu2021just}    & \underline{69.0} \scriptsize{$\pm$3.3} & \underline{77.8} \scriptsize{$\pm$0.3}  \\
            & SD   \cite{pezeshki2021gradient}     & 40.4 \scriptsize{$\pm$2.7}  & 66.5 \scriptsize{$\pm$1.1}  \\
            & Debian \cite{li2022discover} & 33.2 \scriptsize{$\pm$8.0}  & 61.1 \scriptsize{$\pm$2.1}  \\
            & FLAC-B  \cite{sarridis2023flac} & 28.5 \scriptsize{$\pm$4.3}  & 57.3 \scriptsize{$\pm$1.7}  \\
            & \METHOD  & \textbf{84.4} \scriptsize{$\pm$2.2}  & \textbf{89.3} \scriptsize{$\pm$1.3}  \\

        \bottomrule
    \end{tabular} 
\end{table}
\begin{table*}[t]
\centering
\caption{Open-set performance on ImageNet9 in terms of accuracy across 7 test sets.} 
\label{tab:ImageNet9}
 \resizebox{\linewidth}{!}{
\begin{tabular}{lccccccc} 
\toprule
Method & MIXED-NEXT ($\uparrow$) & MIXED-RAND ($\uparrow$) 
& NO-FG ($\downarrow$) & ONLY-BG-B ($\downarrow$) & ONLY-BG-T ($\downarrow$) & ONLY-FG ($\uparrow$) & ORIGINAL ($\uparrow$)
\\ 
\midrule       

Vanilla & 82.66 \scriptsize{$\pm$0.1}	& 85.06 \scriptsize{$\pm$0.0}	
&	64.16 \scriptsize{$\pm$0.1}	&	35.18 \scriptsize{$\pm$0.1}	&	44.74 \scriptsize{$\pm$0.1}	&	\underline{93.12} \scriptsize{$\pm$0.0}	&	97.69\scriptsize{$\pm$0.0}	\\
LfF \cite{nam2020LfF} & 78.70 \scriptsize{$\pm$0.1} & 81.47 \scriptsize{$\pm$0.2} 
& 61.07 \scriptsize{$\pm$0.1} & 34.82 \scriptsize{$\pm$0.2} & 44.46 \scriptsize{$\pm$0.0} & 88.99 \scriptsize{$\pm$0.2} & 94.34 \scriptsize{$\pm$0.2} \\
JTT \cite{liu2021just} & 84.43 \scriptsize{$\pm$0.1} & 86.16 \scriptsize{$\pm$0.5} 
& 61.09 \scriptsize{$\pm$2.0} & 32.04 \scriptsize{$\pm$1.0} & \underline{36.62} \scriptsize{$\pm$4.7} & 92.09 \scriptsize{$\pm$0.5} & 97.71 \scriptsize{$\pm$0.1} \\
Debian \cite{li2022discover} & 83.02 \scriptsize{$\pm$0.4} & 85.64 \scriptsize{$\pm$0.3}
& 64.53 \scriptsize{$\pm$0.4}& 34.45 \scriptsize{$\pm$0.1}& 45.00 \scriptsize{$\pm$0.6}& 93.06 \scriptsize{$\pm$0.1}& \underline{97.89} \scriptsize{$\pm$0.1}\\

SD \cite{pezeshki2021gradient}& \underline{87.56} \scriptsize{$\pm$0.57} & \underline{88.92} \scriptsize{$\pm$0.74} & 62.60 \scriptsize{$\pm$1.05} & 31.42 \scriptsize{$\pm$2.93} & 40.81 \scriptsize{$\pm$3.00} & \textbf{93.71} \scriptsize{$\pm$0.71} & \textbf{98.16} \scriptsize{$\pm$0.06} \\
FLAC-B \cite{sarridis2023flac}& 84.60 \scriptsize{$\pm$0.46} & 86.62 \scriptsize{$\pm$0.45} &\underline{59.84} \scriptsize{$\pm$1.67} & \underline{29.71} \scriptsize{$\pm$0.53} & 40.38 \scriptsize{$\pm$1.28} & 92.72 \scriptsize{$\pm$0.73} & 97.89 \scriptsize{$\pm$0.09} \\
							
\METHOD & \textbf{88.26}\scriptsize{$\pm$0.1}  \normalsize\textcolor{blue}{(+0.70)}& \textbf{89.64}\scriptsize{$\pm$0.2} \normalsize\textcolor{blue}{(+0.72)} 
& \textbf{53.02}\scriptsize{$\pm$0.7} \normalsize\textcolor{blue}{(-6.82)}& \textbf{21.83}\scriptsize{$\pm$0.4} \normalsize\textcolor{blue}{(-7.88)}& \textbf{32.48}\scriptsize{$\pm$0.6} \normalsize\textcolor{blue}{(-4.14)}& 91.90\scriptsize{$\pm$0.4} \normalsize\textcolor{red}{(-1.81)}& 96.92\scriptsize{$\pm$0.2} \normalsize\textcolor{red}{(-1.24)}\\
\bottomrule
\end{tabular}}
\end{table*}

\subsection{Evaluation Protocol}
To assess bias, we primarily use worst-group accuracy (WG Acc.), which measures the accuracy of the least-performing group within a dataset, and average accuracy (Avg. Acc.), which is the mean accuracy across all groups and reflects overall model performance.
The formation of these groups determines whether the evaluation supports a closed- or open-set bias scenario. Accordingly, we implement two distinct evaluation protocols. The first one is designed to align with our approach, focusing on open-set biases, while the second protocol adheres to established evaluation standards in the literature (i.e., closed-set scenario). Furthermore, since \METHOD\ is designed for BU scenarios, we evaluate its performance against the following widely-employed and competitive BU methods: LfF \cite{nam2020LfF}, JTT \cite{liu2021just}, SD \cite{pezeshki2021gradient}, Debian \cite{li2022discover}, and FLAC-B \cite{sarridis2023flac}.
\newline
\textbf{Open-set}. Here, we form groups based on the presence/ absence of the detected open-set bias attributes (i.e., no predefined biases are considered). To achieve that, we use a vanilla model and the potential biases $\mathcal{B}$ extracted through \METHOD\ to identify the subset of tags $\mathcal{B}' \subseteq \mathcal{B}$ that actually introduce bias to the model.
We define bias as occurring when the model exhibits increased accuracy on images that include a particular tag, relative to its overall accuracy across the entire dataset.
After deriving $\mathcal{B}'$ for each class within the datasets, we categorize each group of images belonging to a class into two sub-groups based on whether the samples contain at least one of the biased tags. This results in $2 \times p$ groups. Subsequently, we measure the WG Acc. and Avg. Acc. across all groups. To ensure a fair assessment, the optimal training epoch for each method is selected based on the overall accuracy, without considering any information related to the biases.
An exception to this protocol is ImageNet9. In the original ImageNet9 test set, we observed that for several classes, the subgroups without any biases contain very few samples (fewer than 4), making reliable evaluation difficult. We therefore use the official seven test set variations, which allow for a more comprehensive evaluation of the model's reliance on factors beyond the target object: ORIGINAL, ONLY-BG-B (bounding box of a black object), ONLY-BG-T (bounding box of an inpainted object), NO-FG Black (segmented object removed), ONLY-FG (black background), MIXED-RAND (random background of a random class), and MIXED-NEXT (random background of the next class).
\newline
\textbf{Closed-set}. Here, we follow the protocols suggested by the dataset providers and previous works for predefined biases. In particular, for CelebA we use the accuracy of the underrepresented groups (i.e., bias-conflicting accuracy) and the average accuracy across all groups (i.e., unbiased accuracy) \cite{sarridis2023flac,hong2021bb}. For  Waterbirds, we use the WG Acc. and the Avg. Acc. across all groups. In the case of UrbanCars, we calculate the weighted average accuracy across groups, referred to as In-Distribution Accuracy (I.D. Acc), with weights determined by group representation ratios. Furthermore, I.D. Acc serves as a baseline for assessing accuracy drops related to background (BG Gap), co-occurring objects (CoObj Gap), and both background and co-occurring objects combined (BG+CoObj Gap). 
\subsection{Comparative Analysis}
\label{sec:results}
Table~\ref{tab:openset} presents the open-set performance comparison across the CelebA, Waterbirds, and UrbanCars datasets. For CelebA, \METHOD\ significantly outperforms competing methods by achieving a 66.7\% (+35.2\%) WG accuracy and 81.4\% (+14\%) Avg. accuracy. 
As expected, most existing BU methods exhibit poor performance, as they struggle to handle scenarios with multiple concurrent biases \cite{li2023whac}.
For the Waterbirds, \METHOD\ reaches a WG accuracy of 75.4\% (+10.7\%) and an Avg. accuracy of 87.5\% (+2.3\%), surpassing competing approaches by a notable margin. Similarly, in the UrbanCars dataset, \METHOD\ achieves a WG accuracy of 84.4\% (+15.4\%) and an Avg. accuracy of 89.3\% (+11.5\%). 
For ImageNet9, as shown in Tab.~\ref{tab:ImageNet9}, \METHOD\ achieves accuracy improvements on the test sets with modified backgrounds, MIXED-NEXT and MIXED-RAND (+0.7\% and +0.72). 
\looseness=-1

Similarly, for the sets where background information is suppressed (ONLY-BG-B, ONLY-BG-T, and NO-FG) \METHOD\ achieves improvements of +7.88\%, +4.14\%, and +6.82\%, respectively.
As observed, most competitive methods, including \METHOD, underperform on ONLY-FG compared to the vanilla model. This is likely because the vanilla model can exploit biases present in the foreground (e.g., colors) to increase its accuracy. Finally, there is a 1.24\% drop in performance on the ORIGINAL test set, which aligns with expectations, as \METHOD\ is designed to rely less on shortcuts that boost overall performance.
The top-10 biased tags for each Imagenet9 class are reported in Tab.~\ref{tab:biased_tags}. 
Moreover, Tab.~\ref{tab:biased_tags_supp} reports the top-10 irrelevant tags derived using \METHOD\ for CelebA, Waterbirds, and UrbanCars datasets. For CelebA, \METHOD\ successfully detects the primary bias (i.e., gender) while also revealing other biases associated with facial expressions (smile), shot types (selfies), and clothing items (e.g., dresses, business suits, or ties). In the case of Waterbirds, we observe that, in addition to background elements, waterbirds are predominantly depicted in flight, whereas landbirds are often observed perched on tree branches. Finally, for UrbanCars, as anticipated, the top irrelevant tags correspond to objects that are common in urban/rural environments.
\begin{table}[ht]
\centering
\caption{Top-10 biased tags for each class of ImageNet9.}
\renewcommand{\arraystretch}{0.5} 
\resizebox{\linewidth}{!}{
\begin{tabular}{p{1.5cm}p{5.64cm}} 
\toprule
  Class & Top-10 Irrelevant Tags \\
\midrule

     \multirow{2}{*}{Bird} & perch, sit, branch, tree, tree branch, water, blue, twig, brown, sky \\
\midrule
     \multirow{2}{*}{Carnivore} & stand, grass, stone, lush, fur, walk, lay, enclosure, tree, red \\
\midrule
     \multirow{2}{*}{Dog} & brown, white, stand, black, lush, stare, green, carpet, blanket, bed \\
\midrule
     \multirow{4}{*}{Fish} & sea, swim, water, underwater, aquarium, tank, coral reef, catch, man, yellow \\
\midrule
     \multirow{2}{*}{Insect} & green, plant, break, floor, sit, stem, leaf, close-up, white, yellow \\
\midrule
     \multirow{2}{*}{Instrument} & play, woman, ceiling, table, pillar, cloth, hang, band, tool, hand \\
\midrule
     \multirow{2}{*}{Primate} & tree, sit, branch, tree branch, stare, black, enclosure, brown, stone, log \\
\midrule
     \multirow{2}{*}{Reptile} & floor, stone, green, branch, tree, grass, tree branch, water, lay, sit \\
\midrule
 \multirow{2}{*}{Vehicle} & road, track, red, building, curb, equipment, travel, load, train track, railroad \\
\bottomrule
\end{tabular}
}
\label{tab:biased_tags}
\end{table}

\begin{table}[ht]
\centering
\caption{Top-10 biased tags for each class of CelebA, Waterbirds, and UrbanCars.}
\renewcommand{\arraystretch}{0.5} 
\resizebox{\linewidth}{!}{
\begin{tabular}{p{.1cm}p{1.74cm}p{5.4cm}} 
\toprule
 $\mathcal{D}$ & Class & Top-10 Irrelevant Tags \\
\midrule
\multirow{8}{*}{\centering\rotatebox{90}{CelebA}} 
    & \multirow{2}{*}{Blonde} & woman, dress, pose, smile, girl, beautiful, curl, selfie, actor, carpet \\
\cmidrule(lr){2-3}
    & \multirow{2}{*}{Non-Blonde} & wear, man, shirt, selfie, tie, black, stand, stare, dress shirt, business suit \\
\midrule
\multirow{5}{*}{\centering\rotatebox{90}{Waterbirds}} 
    & \multirow{2}{*}{Landbird} & stand, perch, sit, tree, yellow, branch, floor, brown, forest, green \\
\cmidrule(lr){2-3}
    & \multirow{2}{*}{Waterbird} & water, stand, white, sea, fly, lake, sky, ledge, pond, stone \\
\midrule
\multirow{8}{*}{\centering\rotatebox{90}{UrbanCars}} 
    & \multirow{2}{*}{Country Car} & road, rural, animal, stand, white, jump, cow, lift, bull, highway \\
\cmidrule(lr){2-3}
    & \multirow{2}{*}{Urban Car} & park, house exterior, house, red, building, flip, home, lift, white, jump \\
\bottomrule
\end{tabular}
}
\label{tab:biased_tags_supp}
\end{table}
For the closed-set evaluation, as reported in Tab.~\ref{tab:summary_performance} \METHOD\ achieves +2.7\% (89.7\%) Unbiased and +2.2 (87.1\%) Bias-conflict accuracy compared to the second-best performing method, on CelebA.
On Waterbirds, \METHOD\ attains 93.7\% WG Acc., outperforming all compared BU methods (+5\%), while slightly improving the average accuracy (+0.7\%).
On UrbanCars, \METHOD\ significantly enhances performance by reducing the BG, CoObj, and BG+CoObj gaps to 4.1, 2.4, and 6.7, respectively, representing absolute improvements of 4, 8.1, and 33.4. The full closed-set results are provided in the suppl. material.

\begin{table}[t]
\centering
\caption{Brief closed-set performance comparison between \METHOD\ and competitive BU methods. 
}
\label{tab:summary_performance}
\footnotesize
\resizebox{\linewidth}{!}{
\begin{tabular}{lccc}
\toprule
Dataset & Metric & Best BU Method & \METHOD \\ 
\midrule
\multirow{2}{*}{{CelebA}} & Unbiased & {87.0}\scriptsize{$\pm$0.6} \cite{sarridis2023flac} & \textbf{89.7}\scriptsize{$\pm$0.6} \\ 
& Bias-Conflict & {84.9}\scriptsize{$\pm$2.2} \cite{sarridis2023flac} & \textbf{87.1}\scriptsize{$\pm$1.7} \\
\midrule
\multirow{2}{*}{{Waterbirds}} & WG Acc. & {88.7}\scriptsize{$\pm$0.4} \cite{wu2023discover} & \textbf{93.7}\scriptsize{$\pm$0.4} \\ 
& Avg. Acc. & {93.8}\scriptsize{$\pm$0.7} \cite{wu2023discover} & \textbf{94.5}\scriptsize{$\pm$0.4} \\ 
\midrule
\multirow{3}{*}{UrbanCars} & BG Gap & {8.1} \cite{liu2021just} & \textbf{4.1}\scriptsize{$\pm$0.6} \\ 
& CoObj Gap & {10.5} \cite{li2022discover} & \textbf{2.4}\scriptsize{$\pm$1.4} \\ 
& BG+CoObj Gap & {40.1} \cite{liu2021just} & \textbf{6.7}\scriptsize{$\pm$1.4} \\ 
\bottomrule
\end{tabular}}
\end{table}

\begin{figure}[h]
    \centering
    \begin{subfigure}{\linewidth}
        \centering
        \includegraphics[width=\linewidth,trim={0.5cm 0cm 1cm 0.75cm},clip]{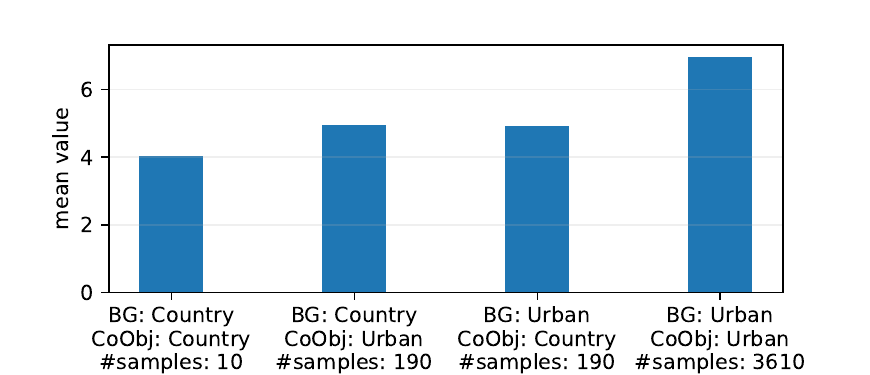}
        \caption{Vanilla}
    \end{subfigure}
    \begin{subfigure}{\linewidth}
        \centering
        \includegraphics[width=\linewidth,trim={0.5cm 0cm 1.cm 0.75cm},clip]{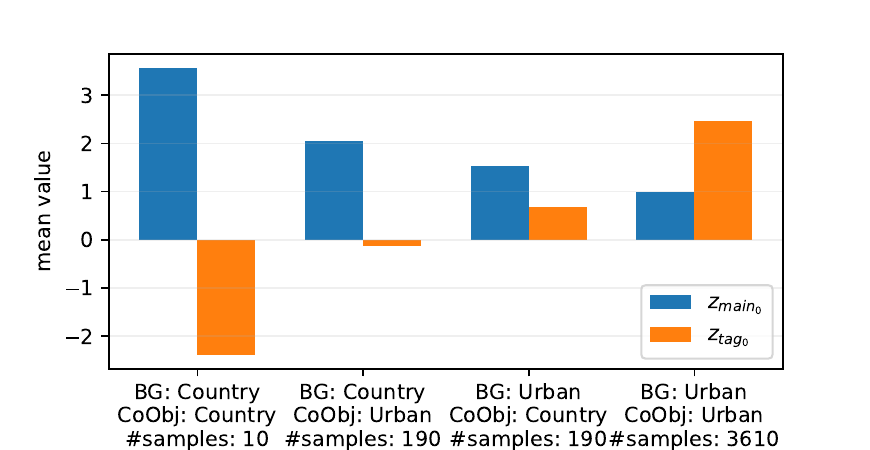}
        \caption{\METHOD}
    \end{subfigure}
    \caption{Logits for UrbanCars training samples belonging to groups defined by the \textit{urban car} class and Background (BG) and Co-occurring Object (CoObj) biases.}
    \label{fig:urbancars_logits}
\end{figure}

In addition, 
by examining the logit spaces,
as shown in Fig.~\ref{fig:urbancars_logits} (a), the vanilla model increases logit values proportionally with the bias, ranging from ~4 to ~7. This is an expected behavior, as the biases are easier to learn than the actual target and thus the model illustrates higher confidence for the bias-aligned samples. This phenomenon is referred to as Gradient Starvation \cite{pezeshki2021gradient} and describes how bias is introduced. Also, it is theoretically shown that methodologies penalizing high logit values can mitigate this phenomenon \cite{pezeshki2021gradient}. As shown in Fig.~\ref{fig:urbancars_logits} (b), the interactions between the main model and the projection layer in the training procedure (defined by Eq.~\eqref{eq:logits_addition} and \eqref{eq:loss_reg}) reduce the main model's logit values proportionally to the data bias.




\subsection{Ablation Analysis}
\label{sec:abl}

An important aspect in \METHOD\ is how the embeddings are calculated. Table~\ref{tab:abl_enc} compares two well-known vision-language models, i.e., OpenCLIP and SigLip, and the text-based model BERT.
\begin{table}[t]
\centering
\caption{Impact of the employed encoders on the \METHOD\ performance. Results pertain to the Waterbirds dataset.}
\label{tab:abl_enc}
\begin{tabular}{lcccc} 
\toprule
Model & WG Acc. & Avg. Acc.\\ 
\midrule 
BERT-L \cite{bert} & 91.7\scriptsize{$\pm$1.3} & 92.8\scriptsize{$\pm$1.0}  \\
SigLip-L-16-256 \cite{zhai2023sigmoid}& 92.7\scriptsize{$\pm$1.2} & 93.5\scriptsize{$\pm$1.0}  \\
OpenCLIP (ViT-L-14) \cite{clip} & 93.7\scriptsize{$\pm$0.4} & 94.5\scriptsize{$\pm$0.4} \\
\bottomrule
\end{tabular}

\end{table}
Furthermore, Tab.~\ref{tab:abl_llm} compares \METHOD's performance using various LLMs. GPT-4o shows the highest performance, with open-source models also performing competitively. For experiments, we use GPT-4o to minimize LLM-related errors. 
Table~\ref{tab:precision_recall} presents the performance of GPT-4o in deriving relevant tags for the Waterbirds, CelebA, and UrbanCars datasets. The ground truth was manually established.

\begin{table}[t]
\centering
\caption{Impact of the employed LLM on the \METHOD\ performance. Results pertain to the Waterbirds dataset.}
\label{tab:abl_llm}
\resizebox{\linewidth}{!}{
\begin{tabular}{lccccc} 
\toprule
Model & \#parameters & WG Acc. & Avg. Acc.\\ 
\midrule   
Qwen2.5 \cite{hui2024qwen2}& 7B & 91.9\scriptsize{$\pm$2.4} & 94.4\scriptsize{$\pm$ 0.5}  \\
Mistral-small \cite{jiang2023mistral}& 22B & 92.9\scriptsize{$\pm$0.4} & 94.9\scriptsize{$\pm$0.2} \\
Gemma2 \cite{team2024gemma}& 9B & 93.5\scriptsize{$\pm$0.6} & 94.5\scriptsize{$\pm$0.6} \\
Llama3.1  \cite{dubey2024llama}& 8B & 93.6\scriptsize{$\pm$0.6} & 94.4\scriptsize{$\pm$0.4}  \\
GPT-4o \cite{openai2023chatgpt4}& $>>$175B & 93.7\scriptsize{$\pm$0.4} & 94.5\scriptsize{$\pm$0.4} \\
\bottomrule
\end{tabular} } 
\end{table}
\begin{table}[t]
\centering
\caption{Performance of GPT-4o.} 
\label{tab:precision_recall}
\begin{tabular}{lcc}
\toprule
Dataset & Precision & Recall \\
\midrule
Waterbirds & 96.1 & 79.0 \\
CelebA & 81.8 & 75.0 \\
UrbanCars & 89.2 & 71.7 \\
\bottomrule
\end{tabular}
\end{table}
Furthermore, we investigate the impact of the different \METHOD\ components. First, as discussed in Sec.~\ref{sec:method}, the logit alignment term in Eq.~\eqref{eq:loss_reg} is crucial for balancing the logits between the two models. Table~\ref{tab:abl_reg} shows the impact of removing this term from the training on the effectiveness of \METHOD.
\begin{table}[t]
\centering
\caption{Impact of Eq.~\eqref{eq:loss_reg} on the \METHOD\ performance. Results pertain to Waterbirds dataset.}
\label{tab:abl_reg}
\begin{tabular}{lcccc} 
\toprule
Type & WG Acc. & Avg. Acc.\\ 
\midrule       
w/o Eq.~\eqref{eq:loss_reg}  & 89.4\scriptsize{$\pm$1.6} & 95.2\scriptsize{$\pm$0.4}  \\
w/ Eq.~\eqref{eq:loss_reg}  & 93.7\scriptsize{$\pm$0.4} & 94.5\scriptsize{$\pm$0.4} \\
\bottomrule
\end{tabular}
\end{table}

In addition, let us report the accuracy of \( g_{\boldsymbol{\phi}}(\cdot) \) on the training samples of the \textit{urban car} class from the UrbanCars dataset, which correlates the target classes with relevant backgrounds and co-occurring objects. As shown in Tab.~\ref{tab:urbancars_acc}, the $g_{\boldsymbol{\phi}}$ shows a clear distinction in its performance: its accuracy increases from 0\% for bias-conflicting samples (i.e., those with a \textit{country} background and co-occurring object) to 98.42\% for bias-aligned samples (i.e., those with an \textit{urban} background and co-occurring object). This result confirms that our tag-based approach produces embeddings that effectively capture bias, providing a foundation for the main model \( f_{\boldsymbol{\theta}}(\cdot) \) to focus on learning unbiased features.

\begin{table}[t]
    \centering
    \caption{\METHOD : $g_{\boldsymbol{\theta}}(\cdot)$ accuracy for UrbanCars training samples belonging to subgroups of \textit{urban car} class with different biases.}
    \begin{tabular}{cccc}
    \toprule
     BG & CoObj &\#samples & $g_{\boldsymbol{\theta}}(\cdot)$ acc. \\ \midrule
      Country & Country & 10 & 00.00\\
      Country & Urban & 190 & 54.21 \\
      Urban & Country & 190 & 67.36\\
      Urban & Urban & 3610 & 98.42\\ \bottomrule
    \end{tabular}
    \label{tab:urbancars_acc}
\end{table}

Furthermore, we investigate the impact of the vocabulary size employed for image tagging. The original vocabulary of RAM has a size of 4585 words. Tab.~\ref{tab:voc_size} reports the performance of \METHOD\ for different portions of the original vocabulary. 
Notably, using 30\% or more of the original vocabulary yields highly effective models. 

\begin{table}[h]
\centering
\caption{Performance of \METHOD\ across varying portions of the original tag vocabulary. Results pertain to the Waterbirds dataset.}
\label{tab:voc_size}
\begin{tabular}{lcccc} 
\toprule
size & WG Acc. & Avg. Acc.\\ 
\midrule       

10\%  & 78.1 & 87.1 \\
20\%  & 79.4 & 89.0 \\
30\%  & 91.1 & 92.0 \\
40\%  & 93.1 & 92.3 \\
50\%  & 93.8 & 93.2 \\
60\%  & 93.2 & 93.5 \\
70\%  & 92.9 & 94.2 \\
80\%  & 93.0 & 94.3 \\
90\%  & 93.6 & 94.5 \\
100\%  & 93.7 & 94.5 \\
\bottomrule
\end{tabular}
\end{table}

\section{Conclusion}

We presented a method to address open-set biases in CV using instance-level descriptive tags. Unlike previous methods relying on predefined biases, \METHOD\ offers a flexible solution for identifying and mitigating unknown biases. It detects visual features that may introduce bias and reduces their impact. 
Extensive experiments show \METHOD\ outperforms existing methods in detecting and mitigating complex, real-world biases. The main requirement is an image tagging model with a comprehensive tag vocabulary and an LLM suited to the application context. 
For example, cloud-deployed models are not ideal for applications where sensitive data is involved, e.g., medical images, images of personal identity documents, etc. 

\section*{Acknowledgments}
This research was supported by the EU Horizon Europe projects
MAMMOth (grant no. 101070285), ELIAS (grant no. 101120237), and ELLIOT (grant no. 101214398).
{
    \small
    \bibliographystyle{ieeenat_fullname}
    \bibliography{main}
}

\input{supp.tex}

\end{document}

%% file: preamble.tex
\usepackage[dvipsnames]{xcolor}


%% file: supp.tex
\clearpage
\setcounter{page}{1}
\maketitlesupplementary
\setcounter{section}{0}

\section{Theoretical Justification}
\subsection{Impact on Gradients}
Let us consider a bias-aligned $\mathbf{x}^{(a)}$ and a bias-conflicting sample $\mathbf{x}^{(c)}$ with targets $y^{(a)}=y^{(c)}=\kappa$. The logits produced by $f_{\boldsymbol{\theta}}$ and $g_{\boldsymbol{\phi}}$ for each class $k\in\mathcal{Y}$, are then $\mathbf{z}^{(a)}_{\text{main}}(k)$, $\mathbf{z}^{(c)}_{\text{main}}(k)$, $\mathbf{z}^{(a)}_{\text{tag}}(k)$, and $\mathbf{z}^{(c)}_{\text{tag}}(k)$, respectively. 
$g_{\boldsymbol{\phi}}$ encodes visual bias, being irrelevant to class $\kappa$, thus we expect $\mathbf{z}^{(a)}_{\text{tag}}(\kappa)\gg\mathbf{z}^{(c)}_{\text{tag}}(\kappa)$. Also, 
bias-aligned samples that contain shortcuts are easier to learn by $f_{\boldsymbol{\theta}}$ and therefore it is expected that $\mathbf{z}^{(a)}_{\text{main}}(\kappa)\ge\mathbf{z}^{(c)}_{\text{main}}(\kappa)$. These imply:
\begin{equation}\label{eq:soft}
\begin{split}
    \mathbf{z}^{(a)}_{\text{main}}(\kappa)+\mathbf{z}^{(a)}_{\text{tag}}(\kappa)&\gg\mathbf{z}^{(c)}_{\text{main}}(\kappa)+\mathbf{z}^{(c)}_{\text{tag}}(\kappa)\Rightarrow\\
    \sigma^{(\kappa)}(\mathbf{z}^{(a)}_{\text{main}}+\mathbf{z}^{(a)}_{\text{tag}})&\gg\sigma^{(\kappa)}(\mathbf{z}^{(c)}_{\text{main}}+\mathbf{z}^{(c)}_{\text{tag}})\Rightarrow\\
    \frac{1}{\sigma^{(\kappa)}(\mathbf{z}^{(a)}_{\text{main}}+\mathbf{z}^{(a)}_{\text{tag}})}&\ll\frac{1}{\sigma^{(\kappa)}(\mathbf{z}^{(c)}_{\text{main}}+\mathbf{z}^{(c)}_{\text{tag}})}
\end{split}
\end{equation}
where $\sigma^{(k)}$ denotes the $k$th class' softmax score.

Additionally, $f_{\boldsymbol{\theta}}$ after a few iterations learns bias-aligned samples (based either on relevant or irrelevant class features) indicating $\argmax_{k\in\mathcal{Y}}\mathbf{z}^{(a)}_{\text{main}}(k)=\kappa$, while $\argmax_{k\in\mathcal{Y}}\mathbf{z}^{(a)}_{\text{tag}}(k)=\kappa$, by definition. Hence, $\sigma^{(\kappa)}(\mathbf{z}^{(a)}_{\text{main}}+\mathbf{z}^{(a)}_{\text{tag}})\approx 1$ leading to diminished gradients $\frac{\partial}{\partial\theta_0}[\sigma^{(\kappa)}(\mathbf{z}^{(a)}_{\text{main}}+\mathbf{z}^{(a)}_{\text{tag}})]\approx 0$. On the other hand, 
$\sigma^{(\kappa)}(\mathbf{z}^{(c)}_{\text{main}}+\mathbf{z}^{(c)}_{\text{tag}})\in (0,1)$ leading to gradients $\frac{\partial}{\partial\theta_0}[\sigma^{(\kappa)}(\mathbf{z}^{(c)}_{\text{main}}+\mathbf{z}^{(c)}_{\text{tag}})]>0$. Hence:
\begin{equation}\label{eq:grad}
    \frac{\partial}{\partial\theta_0}[\sigma^{(\kappa)}(\mathbf{z}^{(a)}_{\text{main}}+\mathbf{z}^{(a)}_{\text{tag}})]\le\frac{\partial}{\partial\theta_0}[\sigma^{(\kappa)}(\mathbf{z}^{(c)}_{\text{main}}+\mathbf{z}^{(c)}_{\text{tag}})]
\end{equation}

Combining \cref{eq:soft} and \cref{eq:grad} we get:
\begin{equation}
    \begin{split}
        \frac{1}{\sigma^{(\kappa)}(\mathbf{z}^{(a)}_{\text{main}}+\mathbf{z}^{(a)}_{\text{tag}})}\cdot \frac{\partial}{\partial\theta_0}[\sigma^{(\kappa)}(\mathbf{z}^{(a)}_{\text{main}}+\mathbf{z}^{(a)}_{\text{tag}})]&\ll \\
        \frac{1}{\sigma^{(\kappa)}(\mathbf{z}^{(c)}_{\text{main}}+\mathbf{z}^{(c)}_{\text{tag}})}\cdot \frac{\partial}{\partial\theta_0}[\sigma^{(\kappa)}(\mathbf{z}^{(c)}_{\text{main}}+\mathbf{z}^{(c)}_{\text{tag}})] &\Rightarrow\\
        \frac{\partial}{\partial\theta_0}[\log{\sigma^{(\kappa)}(\mathbf{z}^{(a)}_{\text{main}}+\mathbf{z}^{(a)}_{\text{tag}})}]&\ll \\
        \frac{\partial}{\partial\theta_0}[\log{\sigma^{(\kappa)}(\mathbf{z}^{(c)}_{\text{main}}+\mathbf{z}^{(c)}_{\text{tag}})}]&\Rightarrow\\
        \frac{\partial}{\partial\theta_0}\bigg[\sum_{k}\mathbbm{1}(a,k)\cdot\log{\sigma^{(k)}(\mathbf{z}^{(a)}_{\text{main}}+\mathbf{z}^{(a)}_{\text{tag}})}\bigg]&\ll \\
        \frac{\partial}{\partial\theta_0}\bigg[\sum_{k}\mathbbm{1}(c,k)\cdot\log{\sigma^{(k)}(\mathbf{z}^{(c)}_{\text{main}}+\mathbf{z}^{(c)}_{\text{tag}})}\bigg]&\Rightarrow\\
        \frac{\partial\mathcal{L}^{cls}(\mathbf{x}^{(a)})}{\partial\theta_0}\ll\frac{\partial\mathcal{L}^{cls}(\mathbf{x}^{(c)})}{\partial\theta_0}
    \end{split}
\end{equation}
where
\begin{equation}
\mathbbm{1}(i,k)=
\begin{cases}
    1  & \text{if } y_i = k \\
    0  & \text{otherwise}
\end{cases}
\end{equation}
which indicates that through our framework, the main model $f_{\boldsymbol{\theta}}$ is enforced to update its weights based mainly on the gradients corresponding to the bias-conflicting samples $\mathbf{x}^{(c)}$. This essentially means that $f_{\boldsymbol{\theta}}$ will eventually ignore the visual cues prevalent in $\mathbf{x}^{(a)}$ samples corresponding to irrelevant information encoded by the tag model $g_{\boldsymbol{\phi}}$.

\subsection{Unbiased Distribution Learning}

Let $p_{tr}(X, Y, B)$ be the training data distribution where $X$, $Y$, $B$ are random variables associated with the data, the label, and the bias attribute, respectively. From Theorem 1 of \cite{hong2021bb} we know that if $P_u$ is a shift of $p_{tr}$ such that the distribution $P_u(Y|B)$ is uniform (and therefore unbiased), then for a model $f_{\boldsymbol{\theta}} = p_u(y|\mathbf{x})$ that estimates the conditional probability over $P_u$, its probability over $p_{tr}$ will be
\begin{equation}
p_{tr}(y|\mathbf{x}) = \frac{\exp\left(\mathbf{z}_{\text{main}}(y) + \log p_{tr}(y|b)\right)}{\sum_{y'}\exp\left(\mathbf{z}_{\text{main}}(y') + \log p_{tr}(y'|b)\right)}
\label{eq:theorem1}
\end{equation}

In the case of \METHOD , let the probability $p_{tr}(y|b)$ be estimated through the projection and classification layers, $\mathbf{z}_{\text{tag}} = f_{\boldsymbol{\theta}_c}(g_{\boldsymbol{\phi}}(\mathbf{e}))$, which use the visual bias embeddings $\mathbf{e}$ (encoding the irrelevant tags) to produce logits $\mathbf{z}_{\text{tag}}$. As a result,
\begin{equation}
\begin{split}
\log p_{tr}(y|b) &= \log\frac{e^{\mathbf{z}_{\text{tag}}(y)}}{\sum_{y''}e^{\mathbf{z}_{\text{tag}}(y'')}}\\
&=\mathbf{z}_{\text{tag}}(y) - \log\left({\sum_{y''}e^{\mathbf{z}_{\text{tag}}(y'')}}\right) \\
&= \mathbf{z}_{\text{tag}}(y) - A(\mathbf{z}_{\text{tag}})
\end{split}
\label{eq:logptrain}
\end{equation}
where $A(\mathbf{z}_{\text{tag}}) = \sum_{y''}e^{\mathbf{z}_{\text{tag}}(y'')}$. Replacing \eqref{eq:logptrain} to \eqref{eq:theorem1} gives
\begin{equation}
\begin{split}
&p_{tr}(y|\mathbf{x}) = \\ =&\frac{\exp\left(\mathbf{z}_{\text{main}}(y) + \mathbf{z}_{\text{tag}}(y) -A(\mathbf{z}_{\text{tag}})\right)}{\sum_{y'}\exp\left(\mathbf{z}_{\text{main}}(y') + \mathbf{z}_{\text{tag}}(y') - A(\mathbf{z}_{\text{tag}})\right)}\\
=& \frac{\exp\left(-A(\mathbf{z}_{\text{tag}})\right)\exp\left(\mathbf{z}_{\text{main}}(y) + \mathbf{z}_{\text{tag}}(y)\right)}{\exp\left(-A(\mathbf{z}_{\text{tag}})\right)\sum_{y'}\exp\left(\mathbf{z}_{\text{main}}(y') + \mathbf{z}_{\text{tag}}(y')\right)}\\
=&\frac{e^{(\mathbf{z}_{\text{main}}(y) + \mathbf{z}_{\text{tag}}(y))}}{\sum_{y'}e^{(\mathbf{z}_{\text{main}}(y') + \mathbf{z}_{\text{tag}}(y')})}
\end{split}
\end{equation}
thus leading to the logit addition of Eq. \eqref{eq:logits_addition}.

For this approach to work, however, one must ensure that models $g_{\boldsymbol{\phi}}$ and $f_{\boldsymbol{\theta}_c}$ are successfully trained to produce logits $\mathbf{z}_{\text{tag}}$ that predict $p_{tr}(y|b)$ and are not dominated by $\mathbf{z}_{\text{main}}$. This is achieved by the logit alignment term of Eq. \eqref{eq:loss_reg} which controls the magnitude differences of $\mathbf{z}_{\text{main}}$ and $\mathbf{z}_{\text{tag}}$.

\section{Experimental Setup}

\subsection{Prompting}
To ensure accurate tag classification, we provide the LLM with a detailed system prompt. Tags are processed in batches of 100, as testing has shown that longer lists can lead to some tags being overlooked. Since relevant tags are significantly fewer than irrelevant ones, we instruct the LLM to return only the relevant tags, allowing us to deduce the irrelevant ones. The exact system prompt is presented in Fig.~\ref{fig:sys_prompt}





\begin{figure}[ht]
\begin{tcolorbox}[colback=white, colframe=black, fonttitle=\bfseries, title= System Prompt]
I will provide you with the name of a target class and a large list of tags. Your task is to evaluate the tags and identify only those directly related to the target class. A tag is considered relevant if it describes or is an essential part of the object associated with the class name. This includes tags that refer to: physical components, defining features, inherent characteristics, and essential behaviors or functions of the object. 

For example, if the target class is ``bee'', tags like ``insect'', ``wing'', and ``buzz'' are relevant because they describe core aspects of what a bee is or does.

Conversely, a tag is irrelevant if it refers to elements that are not an intrinsic part of the object. Irrelevant tags may include: background details, environmental context, colors (unless a defining characteristic), lighting, textures, other objects, or other non-essential contextual elements.

For example, in the case of the class ``bee'', tags like ``sky'', ``flower'', or ``blue'' would be irrelevant, as they describe the environment or background rather than the bee itself.

Please output only the relevant tags in JSON format only (i.e., \{ relevant\_tags: [the list of tags]\}).
\end{tcolorbox}
\caption{LLM system prompt for deriving the relevant tags.}
\label{fig:sys_prompt}
\end{figure}

\section{Two-moon Distribution Experiment}
To demonstrate the capability of \METHOD\ in mitigating biased features, we extend the classic two-moon distribution into a 3-dimensional setting. 
In this scenario, we consider a dataset \(\mathcal{D} = \{(\mathbf{x}^{(i)}, y^{(i)})\}_{i=1}^{N}\), where each input 
\(\mathbf{x}^{(i)} = (x^{(i)}_{1}, 
x^{(i)}_{2}, x^{(i)}_{3})\) 
consists of two relevant features \(x^{(i)}_{1}, x^{(i)}_{2}\) 
and an additional irrelevant feature \(x^{(i)}_{3}\) that introduces bias. The target label \(y^{(i)}\) should be determined based on the features \(x^{(i)}_{1}, x^{(i)}_{2}\), following the typical two-moon structure, while the feature \(x^{(i)}_{3}\) represents the bias (e.g., samples are linearly separable with respect to \(x^{(i)}_{3}\), even though it should not be used for classification).

\begin{figure}[h]
    \centering
    \begin{subfigure}{0.49\linewidth}
        \centering
        \includegraphics[width=\linewidth,trim={0.7cm 0 0cm 0},clip]{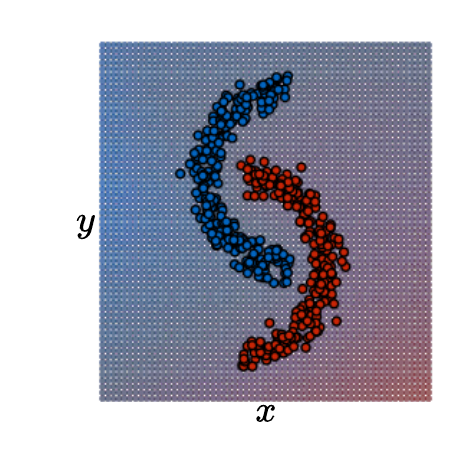}
        \caption{Vanilla outputs on axes $x$ and $y$.}
    \end{subfigure}
    \begin{subfigure}{0.49\linewidth}
        \centering
        \includegraphics[width=\linewidth,trim={0.7cm 0 0cm 0},clip]{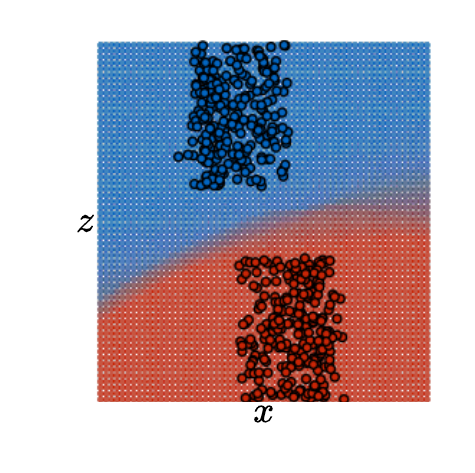}
        \caption{Vanilla outputs on axes $x$ and $z$.}
    \end{subfigure}

    \begin{subfigure}{0.49\linewidth}
        \centering
        \includegraphics[width=\linewidth,trim={0.7cm 0 0cm 0},clip]{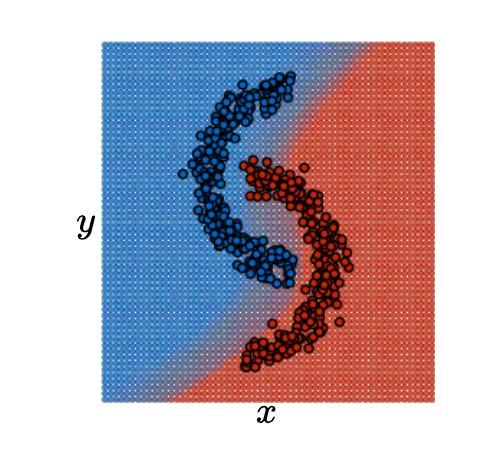}
        \caption{\METHOD\ outputs on axes $x$ and $y$.}
    \end{subfigure}
    \begin{subfigure}{0.49\linewidth}
        \centering
        \includegraphics[width=\linewidth,trim={0.7cm 0 0cm 0},clip]{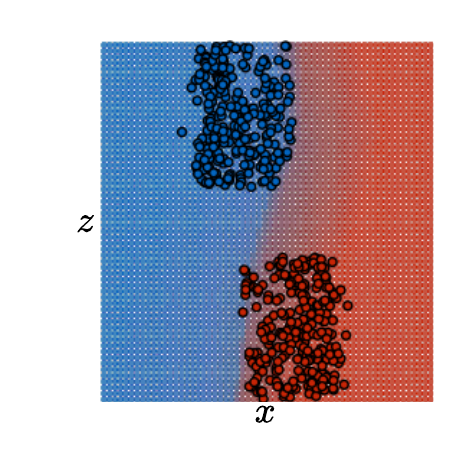}
        \caption{\METHOD\ outputs on axes $x$ and $z$.}
    \end{subfigure}
    \caption{Two-moon problem on 3 dimensions. The distributions are linearly separable on axis $z$ (i.e., $z$ feature introduces bias), while the actual target is to learn the distributions defined by the features $x$ and $y$.}
    \label{fig:twomoons}
\end{figure}

In this setup, the main model \(f_{\boldsymbol{\theta}}(\mathbf{x}^{(i)})\) receives the full input \(\mathbf{x}^{(i)} = (x^{(i)}_{1}, x^{(i)}_{2}, x^{(i)}_{3})\), while the projection layer \(g_{\boldsymbol{\phi}}(x^{(i)}_{3})\) is provided only with the irrelevant feature \(x^{(i)}_{3}\). The projection layer helps the system to explicitly account for this bias by incorporating \(x^{(i)}_{3}\) in a controlled manner, while the main model is encouraged to focus on the relevant features \(x^{(i)}_{1}\) and \(x^{(i)}_{2}\) for classification.
As shown in Fig.~\ref{fig:twomoons}, \METHOD\ enables the main model to effectively learn the underlying two-moon distribution based on the features \(x^{(i)}_{1}, x^{(i)}_{2}\), while the vanilla model relies only on the biased feature \(x^{(i)}_{3}\), treating it as a shortcut that prevents the model from learning the proper distribution.

\section{Comparative Analysis}

In this section, we compare the performance of \METHOD\ with other competitive methods on closed-set (i.e., predefined bias) scenarios. 
The reported results encompass both bias-label aware (BA) and bias-label unaware (BU) methods. However, it should be noted that \METHOD\ can be directly compared only with the BU approaches.
Table~\ref{tab:celeba_full} presents the performance of \METHOD\ on CelebA with \textit{gender} as target and the \textit{BlondHair} as the biased attribute. The performance on unbiased and bias-conflicting samples indicates the ability of \METHOD\ to reduce reliance on spurious correlations between the target and the corresponding biased attributes.
Specifically, \METHOD\ achieves 87.1\% accuracy on bias-conflicting samples, surpassing state-of-the-art BU methods such as FLAC-B \cite{sarridis2023flac} and LfF \cite{nam2020LfF}, while maintaining high unbiased accuracy (89.7\%).
Table~\ref{tab:waterbirds_full} evaluates \METHOD\ on  Waterbirds, a well-known benchmark for spurious correlation problems between bird type and background environment. Notably, \METHOD\ attains 93.7\% Worst-Group accuracy, outperforming all compared BU methods (+5\%), while slightly improving the average accuracy (+0.7\%).
The results on UrbanCars in Tab.~\ref{tab:urbancars_full} further validate the efficacy of \METHOD\ in mitigating multiple biases introduced by co-occurring objects and background features. While the in-distribution accuracy (I.D. Acc) of \METHOD\ is marginally lower than that of LfF \cite{nam2020LfF} or Debian \cite{li2022discover}, it achieves substantial improvements in reducing the BG Gap (4.1) and CoObj Gap (2.4), as well as their combined effects (BG+CoObj Gap: 6.7).

\begin{table}[t]
\begin{center}
 \caption{Bias-conflict and Unbiased accuracy comparison on CelebA with \textit{BlondHair} and \textit{gender} as biased attribute and target, respectively. Bold denotes the best-performing bias label-unaware (BU) method and underlined denotes the best-performing bias label-aware method.}
    \label{tab:celeba_full}
    
\begin{tabular}{lccc}  \toprule
         \multirow{ 3}{*}{Methods}  &  \multirow{ 3}{*}{BU} & \multicolumn{2}{c}{Bias}   \\ \cmidrule(lr){3-4}
       &  &  \multicolumn{2}{c}{BlondHair} \\
       \cmidrule(lr){3-4}
        & &  Unbiased & Bias-conflict  \\ \midrule
         LNL \cite{kim2019lnl} &  \xmark& 80.1\scriptsize{$\pm$0.8}  & 61.2\scriptsize{$\pm$1.5}  \\ 
         
         DI \cite{wang2020DI} &  \xmark& 90.9\scriptsize{$\pm$0.3} & 86.3\scriptsize{$\pm$0.4}  \\ 
         EnD \cite{tartaglione2021end} & \xmark& 86.9\scriptsize{$\pm$1.0} & 76.4\scriptsize{$\pm$1.9} \\ 
         BC-BB \cite{hong2021bb}  & \xmark& \underline{91.4}\scriptsize{$\pm$0.0} & 87.2\scriptsize{$\pm$0.2}  \\ 
         FairKL \cite{barbano2022fairkl} & \xmark & 81.7\scriptsize{$\pm$1.7} & 69.9\scriptsize{$\pm$2.4} \\
         FLAC \cite{sarridis2023flac} &\xmark & 91.2\scriptsize{$\pm$0.3}& \underline{88.7}\scriptsize{$\pm$0.5} \\ \midrule
         LfF \cite{nam2020LfF} &\cmark & 84.2\scriptsize{$\pm$0.3} &  81.2\scriptsize{$\pm$1.4} \\ 
         SoftCon \cite{hong2021bb}  &\cmark & 84.1 & 74.4  \\ 
         FLAC-B \cite{sarridis2023flac}  &\cmark & 87.0\scriptsize{$\pm$0.6} & 84.9\scriptsize{$\pm$2.2}  \\ 
         \METHOD  &\cmark & \textbf{89.7}\scriptsize{$\pm$0.6} & \textbf{87.1}\scriptsize{$\pm$1.7}  \\ \bottomrule 
\end{tabular}
\end{center}
\end{table}
\begin{table}[t]
\centering
\caption{Evaluation on Waterbirds.}
\label{tab:waterbirds_full}
\begin{tabular}{lccccc} 
\toprule
Method & BU & WG Acc. & Avg. Acc.\\ 
\midrule       

GroupDro \cite{sagawa2019distributionally}& \xmark & 90.6\scriptsize{$\pm$1.1} & 91.8\scriptsize{$\pm$0.3} \\
BAdd \cite{sarridis2024badd}& \xmark& \underline{92.9\scriptsize{$\pm$0.3}} & 93.6\scriptsize{$\pm$0.2} \\
DFR \cite{qiu2023simple}& \xmark& \underline{92.9\scriptsize{$\pm$0.2}} & \underline{94.2\scriptsize{$\pm$0.4}} \\
 \midrule
JTT \cite{liu2021just}& \cmark& 86.7\scriptsize{$\pm$1.5} & 93.3\scriptsize{$\pm$0.3} \\
DISC \cite{wu2023discover}& \cmark& 88.7\scriptsize{$\pm$0.4} & 93.8\scriptsize{$\pm$0.7} \\
\METHOD & \cmark & \textbf{93.7}\scriptsize{$\pm$0.4} & \textbf{94.5}\scriptsize{$\pm$0.4}  \\
\bottomrule
\end{tabular}
\end{table}
\begin{table}[t]
\centering
\caption{Evaluation on UrbanCars.}
\label{tab:urbancars_full}

 \resizebox{\linewidth}{!}{
\begin{tabular}{lccccccc} 
\toprule
Method & BU & I.D. Acc & BG Gap & CoObj Gap & BG+CoObj Gap\\ 
\midrule 
GroupDro \cite{sagawa2019distributionally} & \xmark& 91.6 & 10.9  &  3.6 & 16.4 \\
DFR \cite{qiu2023simple}& \xmark & 89.7 & 10.7  &  6.9 & 45.2 \\
LLE \cite{li2023whac}& \xmark & 96.7 & \underline{2.1}  &  2.7 & 5.9 \\
BAdd \cite{sarridis2024badd}& \xmark & 91.0\scriptsize{$\pm$0.7} & 4.3\scriptsize{$\pm$0.4} &  \underline{1.6\scriptsize{$\pm$1.0}} &\underline{3.9\scriptsize{$\pm$0.4}} \\
\midrule 

LfF \cite{nam2020LfF} & \cmark & 97.2 & 11.6  &  18.4 & 63.2 \\
JTT \cite{liu2021just}& \cmark & 95.9 & 8.1  &  13.3 & 40.1 \\
Debian \cite{li2022discover}& \cmark & 98.0 & 14.9  &  10.5 & 69.0 \\
\METHOD & \cmark & 92.8\scriptsize{$\pm$0.8} & \textbf{4.1}\scriptsize{$\pm$0.6} &  \textbf{2.4\scriptsize{$\pm$1.4}} &\textbf{6.7\scriptsize{$\pm$1.4}} \\ 
\bottomrule
\end{tabular}}
\end{table}



\begin{figure}[t]
    \centering
    \includegraphics[width=0.8\linewidth]{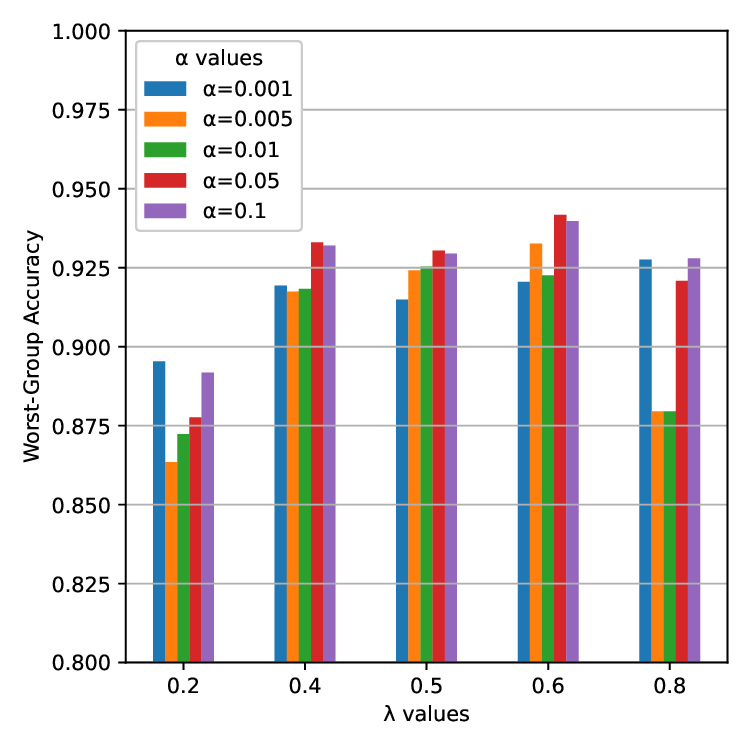}
    \caption{Impact of hyperparameters $\alpha$ and $\lambda$ on the worst-group accuracy (closed-set protocol) on the Waterbirds dataset.}
    \label{fig:hp_waterbirds}
\end{figure}

\section{Ablations \& Hyperparameter Tuning}

Tab.~\ref{tab:abl_emb_type} reports the performance of \METHOD\ when computing individual embeddings for each tag and then averaging them compared to the one computing a single embedding for all tags combined. As one may observe, although both are highly effective, the latter can provide more representative features.
\begin{table}[t]
\centering
\caption{Comparison of \METHOD\ performance on Waterbirds using different approaches for computing embeddings from tags. ``separately'' denotes computing individual embeddings for each tag and then averaging them, while ``collectively'' denotes computing a single embedding for all tags combined.}
\label{tab:abl_emb_type}
\begin{tabular}{lcccc} 
\toprule
Type & WG Acc. & Avg. Acc.\\ 
\midrule       
separately  & 93.0\scriptsize{$\pm$0.2} & 93.6\scriptsize{$\pm$0.2}  \\
collectively  & 93.7\scriptsize{$\pm$0.4} & 94.5\scriptsize{$\pm$0.4} \\
\bottomrule
\end{tabular}
\end{table}

Fig.~\ref{fig:hp_waterbirds} illustrates the impact of different values of hyperparameters $\alpha$ and $\lambda$ on the \METHOD\ performance.  The hyperparameters for all datasets were tuned through the same grid search. Overall, optimal $\lambda$ values are close to 0.5 - we noticed that values $>0.5$ work better for less biased datasets and $<0.5$ for datasets with extreme bias rates. $\alpha$ is a much less sensitive hyperparameter, and any value between 0.001-0.1 does not have a severe impact on the results.

\section{Irrelevant Tags}
Tables~\ref{tab:celeba_irr},\ref{tab:waterbirds_irr},\ref{tab:urbancars_irr}, and \ref{tab:imagenet_irr} report the lists of irrelevant tags for CelebA, Waterbirds, UrbanCars, and ImageNet9 datasets, respectively.
\begin{table*}[h]
\centering
\tiny 
\caption{CelebA irrelevant tags.}
\begin{tabular}{p{0.05\linewidth}p{0.95\linewidth}}
\toprule
\textbf{class} & \textbf{tags}\\
\bottomrule
\end{tabular}

\label{tab:celeba_irr}
\end{table*}

\begin{table*}[ht]
\centering
\tiny 
\caption{Waterbirds irrelevant tags.}
\begin{tabular}{p{0.05\linewidth}p{0.95\linewidth}}
\toprule
\textbf{class} & \textbf{tags}\\
\bottomrule
\end{tabular}

\label{tab:waterbirds_irr}
\end{table*}

\begin{table*}[ht]
\centering
\tiny 
\caption{UrbanCars irrelevant tags.}
\begin{tabular}{p{0.05\linewidth}p{0.95\linewidth}}
\toprule
\textbf{class} & \textbf{tags}\\
\midrule
urban-car & advertisement, alley, apartment, archway, atm, attach, back, backyard, balcony, balustrade, bear, bin, binocular, bird, black, brick building, brush, building, bus stop, bush, cage, canopy, carpet, catch, ceiling, ceiling fan, charge, chase, chimney, city, city skyline, city street, city view, cliff, cloth, collage, corn field, corridor, country lane, countryside, courthouse, crane, crash, creek, crop, crosswalk, curb, cut, dark, destruction, display, dog house, doodle, door, doorway, dune, dust, enclosure, evergreen, exit, explosion, fence, fill, fire truck, fireplace, flip, flood, floor, fuel, garage, garage door, gas, gas station, gravel, green, greenery, gun, hand, hang, hay, head, headlight, hedge, home, hood, horseback, house, house exterior, hurdle, hut, hydrant, ice skating, illuminate, intersection, jockey, jump, lake, lamp, lamp post, lawn, ledge, lift, lighthouse, line, living room, mat, meter, miniature, mountain, narrow, neon light, night, night view, number, office building, out, outdoor, outhouse, overpass, paling, palm tree, park, parking garage, parking lot, parking meter, parking sign, pavement, peak, pen, pergola, phone box, pillar, pine, pine forest, pink, plant, plug, pole, post, pot, power line, pump, rail, railroad, rain, ramp, ravine, red, reindeer, rink, rise building, road sign, row, sale, sculpture, shed, shelter, sign, signal, silo, skate, skyscraper, smoke, snow, snowboarder, snowy, speed limit sign, stable, stair, stall, stare, stick, stone building, stool, stop sign, store, storefront, street corner, street scene, street sign, stunt, tarp, telescope, terrace, toy, track, traffic light, traffic sign, train, train car, train track, trick, trim, tunnel, umbrella, unicorn, urban, vegetation, vineyard, wall, warehouse, warning sign, water, wheat field, white, white house, yak, yard, yellow, zebra crossing \\
\bottomrule
\end{tabular}

\label{tab:urbancars_irr}
\end{table*}

\begin{table*}[ht]
\centering
\tiny 
\caption{ImageNet9 irrelevant tags.}
\begin{tabular}{p{0.05\linewidth}p{0.95\linewidth}}
\toprule
\textbf{class} & \textbf{tags}\\
\bottomrule
\end{tabular}

\label{tab:imagenet_irr}
\end{table*}